\documentclass{article}
\usepackage[utf8]{inputenc}
\usepackage{xcolor}
\usepackage[margin=1in]{geometry}
\usepackage[utf8]{inputenc}
\usepackage{mathtools}
\usepackage{amsmath}
\usepackage{amsthm}
\usepackage{amsfonts}       
\usepackage{nicefrac}       
\usepackage{xcolor}
\usepackage{soul}
\usepackage{dsfont}
\usepackage{natbib}
\usepackage{graphicx}
\usepackage{subfig}
\usepackage{multirow}
\usepackage{algorithm}
\usepackage{algorithmic}
\usepackage{comment}
\newtheorem{theorem}{Theorem}
\usepackage{authblk}

\graphicspath{ {Figures/} }

\title{A Learning-Based Optimal Uncertainty Quantification Method and Its Application to Ballistic Impact Problems}
\author[1]{Xingsheng Sun}
\author[2]{Burigede Liu\footnote{Corresponding author: bl377@cam.ac.uk}}

\affil[1]{University of Kentucky, Lexington, United States}
\affil[2]{University of Cambridge, Cambridge, United Kingdom}
\date{}
\begin{document}

\maketitle
\begin{abstract}
This paper concerns the study of optimal (supremum and infimum) uncertainty bounds for systems where the input (or prior) probability measure is only partially/imperfectly known (e.g., with only statistical moments and/or on a coarse topology) rather than fully specified. Such partial knowledge provides constraints on the input probability measures. The theory of Optimal Uncertainty Quantification allows us to convert the task into a constraint optimization problem where one seeks to compute the least upper/greatest lower bound of the system's output uncertainties by finding the extremal probability measure of the input. Such optimization requires repeated evaluation of the system's performance indicator (input to performance map) and is high-dimensional and non-convex by nature. Therefore, it is difficult to find the optimal uncertainty bounds in practice. In this paper, we examine the use of machine learning, especially deep neural networks, to address the challenge. We achieve this by introducing a neural network classifier to approximate the performance indicator combined with the stochastic gradient descent method to solve the optimization problem. We demonstrate the learning based framework on the uncertainty quantification of the impact of magnesium alloys, which are promising light-weight structural and protective materials. Finally, we show that the approach can be used to construct maps for the performance certificate and safety design in engineering practice. 
\end{abstract}
\section{Introduction}

Engineers and scholars are often faced with scientific applications that are significantly influenced by imperfect knowledge, or uncertainties~\citep{roy2011comprehensive, kidane2012rigorous, adams2012rigorous, kovachki2022multiscale}. In the modeling and computing, these uncertainties can stem from a number of sources such as those due to measurement errors, manufacturing processes, natural material variability, initial and boundary conditions of the system, and lingual information~\citep{lucas2008rigorous, liu2015dynamic}. Furthermore, the physical model itself can also introduce significant uncertainties due to the assumptions of the model and the numerical approximations that are adopted in the simulations~\citep{xiao2016quantifying, draper1995assessment}. The effects of these uncertainties with different sources on the solutions of problems must be estimated and evaluated in order to provide decision makers with predictions and quantitative information about the confidence level at which these predictions can be trusted.  

Uncertainties can be classified as either aleatoric or epistemic (also referred to as irreducible vs. reducible uncertainties or objective vs. subjective uncertainties)~\citep{morgan1990uncertainty, oberkampf2010verification, haimes2005risk}. Aleatoric uncertainties are those inherent to the underlying physical phenomena being studied. Examples include random input excitations and noisy experimental measurements. These inherent variations, given sufficient samples of data, can be characterized through a probability density distribution (PDF). The simplest approach for propagating aleatoric uncertainties is Monte Carlo sampling~\citep{hastings1970monte, mackay1998introduction, liu2018forward}. By contrast, epistemic uncertainties arise due to the lack of knowledge about underlying physical phenomena by the analysts conducting the analysis. Some of these epistemic uncertainties, e.g., model uncertainties and numerical approximation errors, can potentially be reduced by gathering more knowledge through experiments, improved numerical approximation, expert opinion, and higher fidelity physics modeling. Traditionally, epistemic uncertainties are represented by non-PDF methods such as intervals~\citep{jiang2011structural, liu2015probability}, the Dempster–Shafer theory~\citep{shafer1992dempster, jiang2013novel} and fuzzy numbers~\citep{dubois1993fuzzy, haag2010identification}. Aleatoric and epistemic uncertainties are not always easily distinguished during the characterization of input variables and modeling and solution of the system. In this case, the combined uncertainties can be characterized by double-loop or nested sampling methods such as probability boxes (also called $p$-boxes), which provide an envelope of possible PDFs in ranges specified by the available information about both uncertainties~\citep{faes2021engineering, ferson2007experimental}.

This work specifically considers systems that are characterized by a validated physics-based model. By uncertainty quantification (UQ), we mean the determination of probabilities of outcomes in systems whose response is stochastic or uncertain due to the uncertainties in the system inputs or operating conditions. A case in point concerns design certification, i.e., the assessment of the probability that the system will perform safely and within specifications. The design of the system is certified if the probability of failure (PoF) to perform safely is below a prespecified failure tolerance~\citep{liu2021hierarchical, sun2020rigorous, topcu2011rigorous}. However, how to determine the exact solution of the PoF poses a crucial challenge due to the multiple types of uncertainties and lack of uncertainty information. Instead of assuming the specific descriptions of the uncertain variables, it may be more meaningful to just consider the restricted statistical information available for the individual random variables and to compute the optimal upper and lower bounds on the PoF through leveraging all the uncertainty information. In order to compute such bounds, in this paper we employ a recently developed method referred to as Optimal Uncertainty Quantification (OUQ)~\citep{owhadi2013optimal}. This method distinguishes itself from the methods presented above by its ability to consider partial information of input (or prior) probability measures  without the needs of specifying/assuming the full probability measure. In particular, OUQ reformulates the infinite-dimensional optimization problem of computing the supremum (i.e., least upper bound) and infimum (i.e., greatest lower bound) of the PoF in terms of a convex combination of Dirac measures in order to solve a finite-dimensional optimization problem~\citep{winkler1988extreme, winkler1979integral}. The partial information, such as the bounds or statistical moments of the random variables, are then considered constraints in the optimization problem. The OUQ strategy has been used in various applications including design of a thermal hydraulic reactor~\citep{stenger2020optimal}, ballistic impact of aluminum alloys~\citep{kamga2014optimal} and magnesium alloys~\citep{sun2022uncertainty}, rupture of soft collagenous tissues~\citep{balzani2017method}, and sheet forming process~\citep{miska2021efficient}.

The solution to OUQ optimization can be numerically computed thanks to Winkler's theorem~\citep{winkler1988extreme}. This powerful theorem gives the basis for practical calculation of the optimal quantity of interest. In this regard, some numerical methods have been explored such as Semi-Definite-Programming~\citep{lasserre2009moments, betro2000methods}, Mystic framework~\citep{mckerns2012optimal}, and canonical moments~\citep{stenger2020optimal}. However, only derivative-free methods have been employed due to the discontinuous Dirac measures in the objective functions. These methods rapidly reach their limitation as the dimension of the optimization problem increases. As a result, most of the OUQ applications mentioned above are restricted to low-moment constraints, e.g., range or mean of random inputs. In addition, the physics-based forward model might not be computationally cheap and differentiable, which can make the OUQ optimization computationally unfeasible. 

There has been a growing interest in using data driven and machine learning methods in solving physics and engineering problems. In particular, deep neural networks have shown success in solving and approximating the solution operator of partial differential equations~\citep{kovachki2021neural, raissi2019physics}. They have been used as surrogate models in multi-scale modeling~\citep{bhattacharya2022learning, liu2022learning,liu2022learning1}, as well as Bayesian inverse problems~\citep{li2020fourier} to achieve orders of magnitude faster computational speed.   

In this work, we develop a learning-based OUQ framework, to address the challenges raised above in problems involving the finding of the optimal uncertainty bounds for the impact of an elasto-plastic AZ31B magnesium alloy. We introduce the OUQ setting, and the learning based OUQ method
in Section~\ref{sec:method}. In Section~\ref{sec:example}, we proceed to illustrate our framework by means of an application concerned with ballistic impact of AZ31B Mg alloy plates. A conclusion with a summary and short discussion is finally presented in Section~\ref{sec:summary}.
\section{Methodology}
\label{sec:method}
In this section, we start with a brief introduction of the OUQ theory, and the readers are referred to ~\citep{owhadi2013optimal, winkler1979integral, winkler1988extreme} for additional details and further references. We then proceed to introduce the learning based framework, with detailed remarks on our approach and its alternatives. 
  
\subsection{Optimal uncertainty quantification}

We are concerned with a system whose performance is described by a known response function
\begin{equation}
    Y = F(X) ,
\end{equation}
from one compact measurable space $\mathcal{X} \subseteq \mathbb{R}^m$ of inputs to a second measurable space $\mathcal{Y} \subseteq \mathbb{R}^n$ of outputs. $X \equiv (x_1,\dots,x_m)$ are $m$ real-valued random variables, expressing imperfectly known or uncertain properties of the system and $Y\equiv (y_1,\dots,y_n)$ are $n$ real-valued random variables, or performance measures. The input variables are generated randomly according to an unknown random variable $X$ with values in $\mathcal{X}$ according to a law $\mathbb{P} \in M(\mathcal{X})$, where $M(\mathcal{X})$ is the set of all probability measures supported on $\mathcal{X}$. The design specifications require that $Y$ remain that $Y \in \mathcal{Y}_\text{a}$ for some \emph{admissible} set $\mathcal{Y}_\text{a} \subseteq \mathcal{Y}$. Thus the design fails if $Y \in \mathcal{Y}_\text{c}$ where $\mathcal{Y}_\text{c} = \mathcal{Y} \backslash \mathcal{Y}_\text{a}$ is the \emph{inadmissible} set. Ideally, we would like the support of the probability measure associated to $Y$ to be contained within $\mathcal{Y}_\text{a}$, i.e.,
\begin{equation}
	\mathbb{P}[Y \in \mathcal{Y}_\text{a}] = 1.
\end{equation} 
Systems satisfying this condition can be certified with complete certainty. However, this absolute guarantee of safe performance may be unattainable, e.g., if $\mathbb{P}$ lacks compact support, or is prohibitively expensive \textcolor{red}. In these cases, we may relax the condition of certification. Let $\epsilon \in [0,~1]$ denote the greatest acceptable probability of failure (PoF). Then we say that the system is safe if 
\begin{equation}
\mathbb{P}[Y \in \mathcal{Y}_\text{c}] \leq \epsilon,
\label{eq:safe}
\end{equation} 
and the system is unsafe if 
\begin{equation}
\mathbb{P}[Y \in \mathcal{Y}_\text{c}] > \epsilon.
\label{eq:unsafe}
\end{equation} 

However, due to the lack of information, the exact probability measure $\mathbb{P}$ may be unknown. To this end, we define a subset
\begin{equation}
    \mathcal{A} \subseteq \big\{\mu~|~\mu \in M(\mathcal{X}) \big\}
    \label{eq:info}
\end{equation}
that encodes all the information that we have about the probability measure of the random variables $\mathbb{P}$. This information may come from experimental data, lower-level simulations or expert opinions. Notably, $\mathbb{P} \in \mathcal{A}$ and some admissible scenarios in $\mathcal{A}$ may be safe (i.e., $\mu [Y \in \mathcal{Y}_\text{c}] \leq \epsilon$), whereas other admissible scenarios may be unsafe (i.e., $\mu [Y \in \mathcal{Y}_\text{c}] > \epsilon$). Now observe that, given such an information/assumptions set $\mathcal{A}$, there exist upper and lower bounds on $\mathbb{P}[Y \in \mathcal{Y}_\text{c}]$ corresponding to the scenarios compatible with assumptions, i.e., the values $U(\mathcal{A})$ and $L(\mathcal{A})$ of the optimization problems
\begin{subequations}
\begin{equation}
  	U(\mathcal{A}) := \sup_{\mu \in \mathcal{A}} \mu [Y \in \mathcal{Y}_\text{c}],
\end{equation}    
\begin{equation}
  	L(\mathcal{A}) := \inf_{\mu \in \mathcal{A}} \mu [Y \in \mathcal{Y}_\text{c}].
\end{equation}
\label{eq:bounds}
\end{subequations}
Assume that $U(\mathcal{A}) $ and $L(\mathcal{A})$ can indeed be computed on demand. Now, since $\mathbb{P} \in \mathcal{A}$, it follows that 
\begin{equation}
	L(\mathcal{A}) \leq \mathbb{P}[Y \in \mathcal{Y}_\text{c}] \leq U(\mathcal{A}).
\end{equation}
The upper bound $U(\mathcal{A})$ is \emph{optimal} in the sense that for all $\mu \in \mathcal{A}$, we have $\mu [Y \in \mathcal{Y}_\text{c}] \leq U(\mathcal{A})$, and if $U'<U(\mathcal{A})$, there exists $\mu \in \mathcal{A}$ such that $U'< \mu [Y \in \mathcal{Y}_\text{c}] \leq U(\mathcal{A})$. Similar conclusions apply for the lower bound $L(\mathcal{A})$.

\begin{figure}[!ht]
\centering
\includegraphics[width=3.0in]{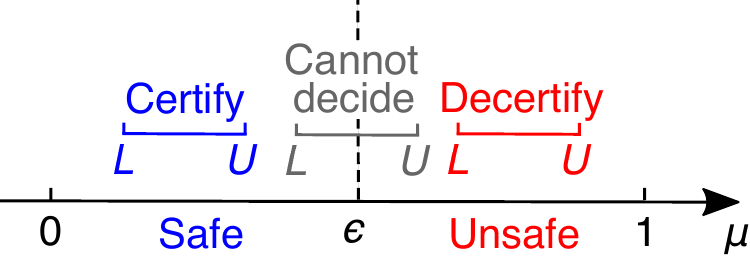}
\caption{Certification process providing a rigorous certification criterion whose outcomes
are of three types: ``certify", ``decertify", and ``cannot decide".}
\label{fig:certify}
\end{figure}

The bounds $U(\mathcal{A})$ and $L(\mathcal{A})$ defined in Eqn.~(\ref{eq:bounds}) can be used to construct a solution to the certification problem. Provided that the information set $\mathcal{A}$ is valid (in the sense that $\mathbb{P} \in \mathcal{A}$), then if $U(\mathcal{A}) \leq \epsilon$, then the system is provably safe; if $\epsilon < L(\mathcal{A})$, then the system is provably unsafe; and if $L(\mathcal{A}) \leq \epsilon < U(\mathcal{A})$, then the safety of the system cannot be decided due to lack of information. The corresponding certification process and its optimality are illustrated in Fig.~\ref{fig:certify}. Evidently, the tighter the bounds $U(\mathcal{A})$ and $L(\mathcal{A})$ the more economical the certification. However, increasing tightness comes at increasing computational expense, which sets forth a fundamental trade-off between economy of certification and computability.

\subsection{OUQ as an optimization problem}
Our goal is to compute the optimal probability $U(\mathcal{A})$ and $L(\mathcal{A})$, as defined in Eqn.~(\ref{eq:bounds}). In general, the OUQ problem is an infinite dimensional optimization problem and is computationally intractable. To this end, we follow Winkler~\citep{winkler1979integral,winkler1988extreme} and show that for the constraint set $\mathcal{A}$ that is of practical interest in uncertainty quantification, the infinite-dimensional optimization problem can be reduced to a finite dimensional optimization problem. In this section, we list the main theorem from Winkler~\citep{winkler1988extreme}, as well as its application to the OUQ problem. We refer the readers to~\citep{owhadi2013optimal} for further references. 

For the sake of brevity, we shall restrict our attention to $U(\mathcal{A})$ as defined in Eqn.~(\ref{eq:bounds}), and the treatment for $L(\mathcal{A})$ follows a similar manner. First note the set of all probability measures supported on $\mathcal{X}$, $M(\mathcal{X})$, is convex. Indeed, given two arbitrary measures $\mu_1, \mu_2 \in M(\mathcal{X})$, the convex combination $\mu_3 = \alpha \mu_1 + (1-\alpha) \mu_2, \alpha \in (0,1)$ is also a probability measure in $M(\mathcal{X})$. Furthermore, it is known that the extremal points of $M(\mathcal{X})$ are Dirac measures of the form~\citep{rudin1991functional}
\begin{equation}
\delta_z(\mathcal{Z}) =   \begin{cases}
      1 & \text{if $z \in \mathcal{Z}$},\\
      0 & \text{otherwise}.
      \end{cases}    
\end{equation}
\begin{theorem}[Winkler's theorem]
\label{Winkler}
Fix K measurable functions $f_i: \mathcal{X} \mapsto \mathbb{R}$, and real values $c_i \in \mathbb{R}$. Consider the set
\begin{equation*}
\mathcal{A} = \{\mu \in \mathcal{M}(\mathcal{X}): f_i~\text{is $\mu$-integrable} \ and \int_{\mathcal{X}}f_id\mu = c_i, \ i \in {1,..,K} \}.  
\end{equation*}
Then $\mathcal{A}$ is convex and its extremal points $\textbf{ex} \ \mathcal{A}$ are given by
\begin{align}
\textbf{ex} \ \mathcal{A} = \bigg\{\mu \in \mathcal{A}: \mu &= \sum_{i=1}^{Q}t_k \cdot \delta_X{(\mathcal{X}_i)}, \mathcal{X}_i=\{X_i: X_i \in \mathcal{X}\},\\ &t_k > 0, \ \sum_{i=1}^{Q} t_i = 1 , \ 1\leq Q \leq K+1\bigg\}.
\end{align}
\end{theorem}
Theorem \ref{Winkler} provides a path to reduce the infinite dimensional optimization problem defined in Eqn.~(\ref{eq:bounds}) to a finite dimensional optimization problem on the extremal points $\textbf{ex}~\mathcal{A}$. 

To see this, first recall that a Choquet type integral representation formula can be derived for probability measures~\citep{winkler1988extreme}. Specifically, for every probability measure in the constrained set $\mu \in \mathcal{A}$ defined in Theorem~\ref{Winkler}, there is a probability measure $p$ supported on $\bf{ex}~\mathcal{A}$, such that  
\begin{equation}
\mu(Y \in \mathcal{Y}_c) = \int_{\nu \in \textbf{ex} \mathcal{A}} \nu(Y \in \mathcal{Y}_c) dp(\nu).
\end{equation}
Subsequently, 
\begin{equation}
\sup\{\mu(Y \in \mathcal{Y}_c): \mu \in \mathcal{A}\} = \sup\bigg\{\int_{\textbf{ex} \mathcal{A}} \nu(B) dp(\nu) \bigg\} = \sup\{\nu(Y \in \mathcal{Y}_c): \nu \in \textbf{ex} \mathcal{A} \}.
\end{equation}
Therefore, the calculation of $U(\mathcal{A})$ can be reduced into an optimization problem of finding the supremum of extremal points in the constraint set $\mathcal{A}$, where
\begin{equation} \label{eq:optimization}
    \begin{aligned}
    &U(\mathcal{A}) = \max_{\substack{X_i\in{\mathcal{X}},i=1,...,K+1 \\  t_i \in [0,1], i= 1,...,K+1}} \sum_{i=1}^{K+1}t_i\delta_{F(X_i)}(\mathcal{Y}_c),\\
    &\text{subject to} \ \ 
    \sum_{i=1}^{K+1}t_i = 1, \ \ \sum_{i=1}^{K+1}t_i \cdot f_j(X_i) = c_j , \ \ j = 1,...,K.
    \end{aligned}
\end{equation}

In addition, we note for the special case where the admissible set contains no constraint, the optimization problem reduced into optimizing a single Dirac measure,
\begin{equation}
U(\mathcal{A}) = \max_{X \in \mathcal{X}}  \delta_{F(X)}(\mathcal{Y}_c),
\end{equation}
which corresponds to finding the worst case scenario.
Furthermore, Theorem~\ref{Winkler} provides a way of constraining the input/prior distribution. As an example, the mean constraint can be imposed by setting $f = X$, while the higher-order moment constraints can be  imposed by setting $f = X^k$, where $k$ is the order of moment. We shall demonstrate this further in the example studies in Section \ref{sec:example}.  

\subsection{Learning based OUQ} \label{sec:OUQ_net}
The implementation of the optimization problem defined in Eqn.~(\ref{eq:optimization}) requires repetitive evaluation of the the performance indicator: 
\begin{equation} \label{eq:NN approx}
\mathcal{D}: X \mapsto \delta_{F(X)}(\mathcal{Y}_c),  
\end{equation}
which is a composition of the system response function $Y = F(X)$ and the Dirac measure $\delta_Y(\mathcal{Y}_c)$. Direct evaluation of the system's response function can become very expensive for high-dimensional, non-linear systems. Our idea is to \textit{learn the performance indicator} from data using a deep neural network and by utilizing \textit{data generated by solutions of the system's response function} over various inputs sampled from an appropriate probability measure defined in the space of input $\mathcal{X}$. To do so, we observe that the output of the performance indicator $\mathcal{D}$ can only take the values from the set $\{0,1\}$, hence the problem closely resembles the classification problem in machine learning~\citep{osisanwo2017supervised}. To that end, we consider a neural network approximation of $\mathcal{D}_{NN}: \mathcal{X} \times \mathbb{R}^{n_\theta} \mapsto \ \{0,1\}$ with parameterization $\theta \in \mathbb{R}^{n_{\theta}}$, such that
\begin{equation} 
\mathcal{D}_{NN}(X;\theta) \approx \mathcal{D}(X).  
\end{equation}
The proposed approach can then be finished in three steps: 
\begin{enumerate}
    \item Data collection: construction of the data set $\{X,Y\}$
    \item Training: For an admissible set $\mathcal{Y}_c$, evaluate and construct the training data set $\{X, \delta_Y(\mathcal{Y}_c)\}$. Then train the neural network $\mathcal{D}_{NN}$ using the training data set.  
    \item Optimization: Optimize Eqn.~(\ref{eq:optimization}) with the surrogate $\mathcal{D}_{NN}$ as an approximation of $\mathcal{D}$. 
\end{enumerate}
We proceed with a series of comments on the learning-based OUQ approach as well as its alternatives. 

\subsubsection*{Data collection}

As opposed to the classical UQ approach, learning based OUQ do not require explicit knowledge on the system's response function $F$. Rather, it requires data in the form of $\{X, Y\}$ sampled from the input space $\mathcal{X}$. A crucial issue in generating data is to balance the cost of generating the data with the need to sample sufficient input to provide an accurate enough approximation for the input encountered in the optimization problem. This leads to the question
of identifying an optimal sampling distribution of input. This remains an active area of research. 

In this study, we use Latin Hypercube Sampling (LHS) to sample the input data from the input space $\mathcal{X} \subset \mathbb{R}^m$, where $\mathcal{X}$ is first divided into $L$ equal-volume subspaces $\mathcal{X}_l \subset {\mathcal{X}}: l = (1,...,L)$. The training input data $\{X\}$ is obtained by sample $N_L$ of inputs $X$ from each subspace $\mathcal{X}_l$ with a uniform probability measure supported on $\mathcal{X}_l$. The total amount of training data is therefore $N_d = L\times N_L$.

\subsubsection*{Neural network architecture and loss function} 
We note the output of the performance indicator $\mathcal{D}$ can only take a value zero or one. In the $D_{NN}$ approximation, we take a standard fully connected neural network architecture with scaled exponential linear unit (SELU)~\citep{klambauer2017self} activation function, and add a Sigmoid activation function~\citep{Goodfellow-et-al-2016} to its last hidden layer so that $\mathcal{D}_{NN}(X;\theta)$ is constrained to $[0,1]$. During the training process, we measure the train and test error with the Binary Cross Entropy (BCE) loss function~\citep{ruby2020binary}, where we define 
\begin{equation}
\text{error} = \sum_{i=1}^{N}\Bigl (Y_i\text{log}(\mathcal{D}_{NN}(X_i;\theta))+(1-Y_i)\text{log}(1-\mathcal{D}_{NN}(X_i;\theta)) \Bigr ). 
\end{equation}
The proposed neural network architecture trained with the BCE loss is very powerful in approximating $\mathcal{D}$, and we demonstrate this further in the next section.

\subsubsection*{OUQ optimization}
The computation of optimal probability bounds as defined in Eqn.~(\ref{eq:optimization}) is a high-dimensional, non-convex optimization problem with multiple constraints. Indeed, the dimension of the optimization problem grows linearly with the input dimension $m$ and the number of moment constraints $K$. Furthermore, for practical problems, it is most of the time difficult or impossible to compute the gradient $\nabla_X{Y}$, hence classically the OUQ problem has to be solved using non-gradient based optimization methods such as genetic algorithms or nested sampling method~\citep{mirjalili2019genetic,skilling2006nested}. These methods require extremely long time to converge and there is in general no guarantee that it converges to the global optimum. 

On the other hand, using a neural network surrogate model to approximate the performance indicator will allow us orders of magnitude faster evaluation of the performance indicator ${D}_{NN}$. Additionally, it is possible to approximate the gradient $\nabla_X \mathcal{D}_{NN}$, which allows us to use the gradient based optimization methods together with penalty methods to impose the moment constraints. Let $\lambda = (\lambda_0, \lambda_1,...,\lambda_K) \in \mathbb{R}^+$ denotes the $K+1$ dimensional penalty coefficients, we define the penalized OUQ loss function $\mathcal{L}_{UQ}$ as
\begin{equation}
\mathcal{L}_{UQ} = \sum_{i=1}^{K+1} \Bigl ( t_i \mathcal{D}_{NN}(X_i;\theta)  \Bigr ) + \lambda_0\Bigl ((\sum_{i=1}^{K+1}t_i)-1\Bigr )^2 +\sum_{j=1}^{K}\Bigl(\lambda_j \bigl(\sum_{i=1}^{K+1}t_if_j(X_i)-c_j\bigr)^2 \Bigr).   
\end{equation}
The optimization is then conducted using the ADAM method~\citep{kingma2014adam}, which belongs to the class of optimizers that utilizes stochastic gradient descent. We highlight that, although the ADAM algorithm provides a much faster convergence rate, there is no guarantee that the solution will converge to the global optimum, especially in the case of high-dimensional non-convex optimization problems. Therefore, for all studies considered in this paper, the corresponding OUQ calculation is repeated roughly $50$ times with random initialization, and we choose the optimal (maximum/minimum) value from the result sets.  
\section{Numerical Examples}
\label{sec:example}

We proceed to illustrate the learning-based OUQ framework described in the foregoing by means of an application concerned with the ballistic impact of an AZ31B Mg alloy plate, Fig.~\ref{fig:setup}(a). We assume the design specification to be a maximum allowable backface deflection of the plate, Fig.~\ref{fig:setup}(b). We say that the system is safe if the maximum backface deflection is less than a given threshold. Otherwise, the design will fail. We further assume that all uncertainty arises from an imperfect characterization of the constitutive response of the plate. As a simple scenario, we assume that, under the conditions of interest, the plate is well described by the Johnson-Cook model~\citep{johnson1983constitutive}, but the model parameters are uncertain. Specifically, they are allowed to vary over certain ranges in order to cover the experimental data. We also know partial information on the input probability measure. We specifically consider the cases where such information is given in the form of statistical moments. For simplicity, the projectile is assumed to be rigid and uncertainty-free. 

\subsection{Material modeling}

We assume that the constitutive behavior of the plate is characterized by an appropriately calibrated Johnson-Cook plasticity model~\citep{johnson1983constitutive},
\begin{equation}
    \sigma \big( \epsilon_p, \dot{\epsilon}_p, T \big)
    =
    \big[A + B \epsilon_p^n \big]
    \big[1 + C \ln \dot{\epsilon}_p^* \big]
    \big[1 - {T ^*} ^m \big],
\label{eq:johnsoncook}
\end{equation}
where $\sigma$ is the true Mises stress, $\epsilon_p$ is the equivalent plastic strain, $\dot{\epsilon}_p$ is the plastic strain rate, and $T$ is the temperature. The normalized plastic strain rate $\dot{\epsilon}_p^*$ and temperature $T^*$ are defined as
\begin{equation}
    \dot{\epsilon}_p^* := \frac{\dot{\epsilon}_p}{\dot{\epsilon}_{p0}},
\label{eq:estar}
\end{equation}
and
\begin{equation}
    T^* := \frac{T - T_0}{T_m - T_0},
\label{eq:tstar}
\end{equation}
respectively, where $\dot{\epsilon}_{p0}$ is a reference strain rate, $T_0$ is a reference temperature and $T_m$ is the melting temperature. The model parameters are: $A$, the yield stress; $B$, the strain-hardening modulus; $n$, the strain-hardening exponent; $C$, the strengthening coefficient of strain rate; and $m$, the thermal-softening exponent.

\begin{table}[!ht]
\centering
\caption{Lower and upper bounds of Johnson-Cook parameters of AZ31B Mg alloy~\citep{hasenpouth2010tensile}.}
\begin{tabular}{l l l}
\hline
\hline
Parameter & ~\  Lower bound  &  ~\ Upper bound  \\
\hline
$A$ (MPa) &     $200.372$~~~ &     $249.970$~~~ \\
$B$ (MPa) &     $150.682$~~~ &     $186.010$~~~ \\
$n$       &       $0.160$~~~ &       $0.324$~~~ \\
$C$       &       $0.012$~~~ &       $0.014$~~~ \\
$m$       &       $1.523$~~~ &       $1.577$~~~ \\
\hline
\end{tabular}
\label{tab:randparam}
\end{table}

\subsection{Problem setup} 
We choose our system's performance measure $Y$, to be the maximum back face deflection of the plate after the impact $Y=y_r \in \mathbb{R}$. We set a maximum backface deflection threshold, $Y_T \in \mathbb{R}$, and consider the system safe if the $y_r \in \mathcal{Y}_a = [0,Y_T)$, and unsafe if $y_r \in \mathcal{Y}_c = [Y_T, +\infty)$. Furthermore, we regard the set $\{A, B, n, C, m\}$ of Johnson-Cook parameters as the main source of uncertainty in the analysis. The bounds of these input uncertain parameters are  tabulated in Table~\ref{tab:randparam}, which are determined by experimental characterization with $95$\% confidence intervals~\citep{hasenpouth2010tensile}. In what follows, we normalize the Johnson-Cook parameters into the range $[0,1]$ using the min-max feature scaling and the corresponding fixed bounds in Table~\ref{tab:randparam}. As a result, the random inputs in the calculations are the normalized Johnson-Cook parameters represented by $X\equiv \{\bar{A}, \bar{B}, \bar{n}, \bar{C}, \bar{m}\} \in [0,1]^5$. We therefore aim to estimate the probability of failure (PoF) $\mathbb{P}[Y \in \mathcal{Y}_c]$, from uncertainties in inputs $X$. 

\begin{figure}[!ht]
\centering
\includegraphics[width=1.0\textwidth]{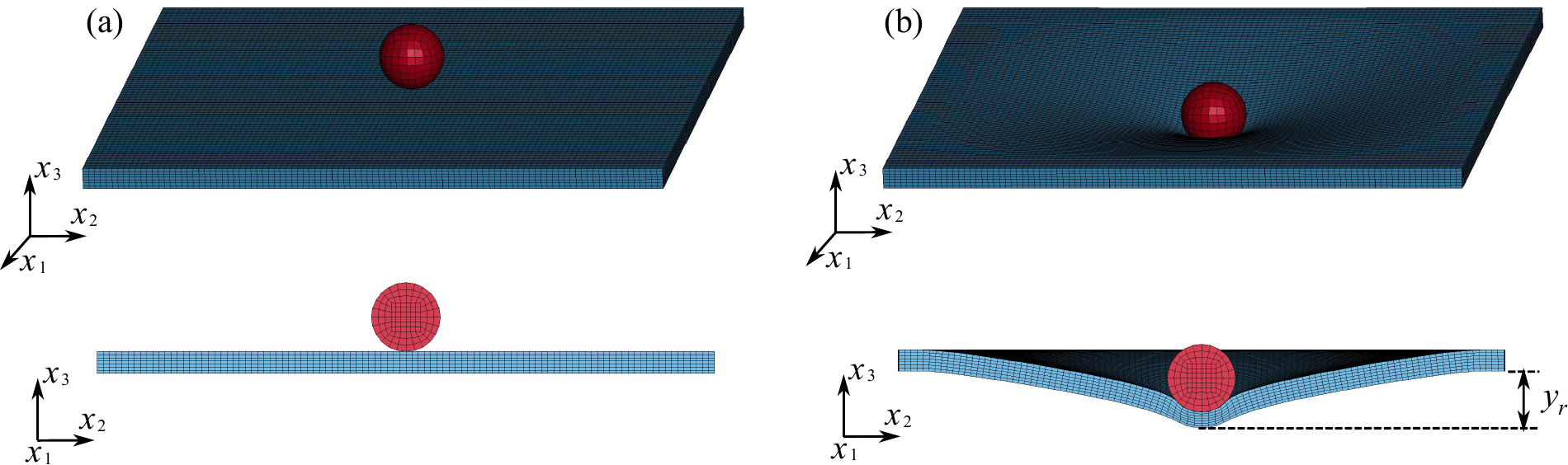}
\caption{Schematic illustration of Mg plate struck by a spherical steel projectile at a ballistic speed. (a) Initial setup. (b) Performance measure with maximum backface deflection labeled as $y_r$. In each subfigure, the top figure shows a perspective view of the projectile/plate system, and the bottom figure shows the view of the middle $x_2$-$x_3$ cross-section.}
\label{fig:setup}
\end{figure}

We use finite element software LS-DYNA~\citep{hallquist2007ls} to solve the impact problem. A schematic of the finite-element model is shown in Fig.~\ref{fig:setup}. The diameter of the projectile is $1.12$~cm, and the size of the plate is $10\times10\times0.35$~cm. The attack velocity is $200$~m/s with normal impact. The backface nodes of the target near the edges are fully constrained to prevent displacement in all directions. The projectile is resolved using $864$ elements, while the number of elements for the plate is $70,000$. All the elements are linear hex, single point integration with careful hourglass control. The time-step size is adaptive and determined by the critical size of elements, with all simulations running for $500.0~\mu\text{s}$ before termination. This simulation duration is sufficiently long to allow for the rebound and separation of the projectile from the plate in all the calculations. The calculations are adiabatic with the initial temperature set at room temperature. The equation-of-state, which controls the volumetric response of the material, is assumed to be of the Gruneisen type. For simplicity, the projectile is assumed to be rigid and uncertainty-free. All other material parameters are fixed and listed in Table~\ref{tab:fixedparam}.

\begin{table}[!ht]
\centering
\caption{Fixed material parameters used in the LS-DYNA simulation.}
\begin{tabular}{l l l l l}
\hline
\hline
& Parameter & Value & Unit  & Source                \\
\hline
\multirow{10}{*}{Target (magnesium)}
& Mass density & $1.77$ & g/$\text{cm}^3$ & - \\
& Young's modulus & $45.0$ & GPa & - \\
& Poisson's ratio & $0.35$ & - & - \\
& Specific heat & $1.04$ & J/(K$\cdot$g) & \cite{lee2013thermal} \\
& Gruneisen intercept & $4520.0$ & m/s & \cite{feng2017numerical} \\
& Gruneisen gamma & $1.54$ & -  & \cite{feng2017numerical}  \\
& Gruneisen slope & $1.242$ & -  & \cite{feng2017numerical} \\
& Reference strain rate           &     $0.001$             & $\text{s}^{-1}$ & \cite{hasenpouth2010tensile}       \\
& Reference temperature           &     $298.0$             & K & \cite{hasenpouth2010tensile}                      \\
& Reference melting temperature           &     $905.0$             & K & \cite{hasenpouth2010tensile}                      \\
\hline
\multirow{3}{*}{Projectile (steel)}
& Mass density & $7.83$ & g/$\text{cm}^3$ & - \\
& Young's modulus & $210.0$ & GPa & - \\
& Poisson's ratio & $0.30$ & - & - \\
\hline
\end{tabular}
\label{tab:fixedparam}
\end{table}

\subsection{Learning the performance indicator} 
We follow the learning-based OUQ method as detailed in Section~\ref{sec:OUQ_net}. A total of 2000 input-output pairs $\{X,Y\}$ are generated using the Latin Hypercube Sampling method with $L = 2000$ and $N_L = 1$. 
We then use this data to compute the performance indicator $\mathcal{D}(X)$ for a variety of $Y_T \in [0.9, 1.3]$~cm. We subsequently train a series of neural networks with the architecture described in Section \ref{sec:OUQ_net}. For each $Y_T$, we use a total of $1500$ samples to train and the remaining $500$ to test the learned indicator. In all cases, neural network consists of $4$ intermediate layers with $200$ nodes per layer and are trained using the ADAM~\citep{kingma2014adam} method.

\begin{figure}[!ht]
\centering
\includegraphics[width=1.0\textwidth]{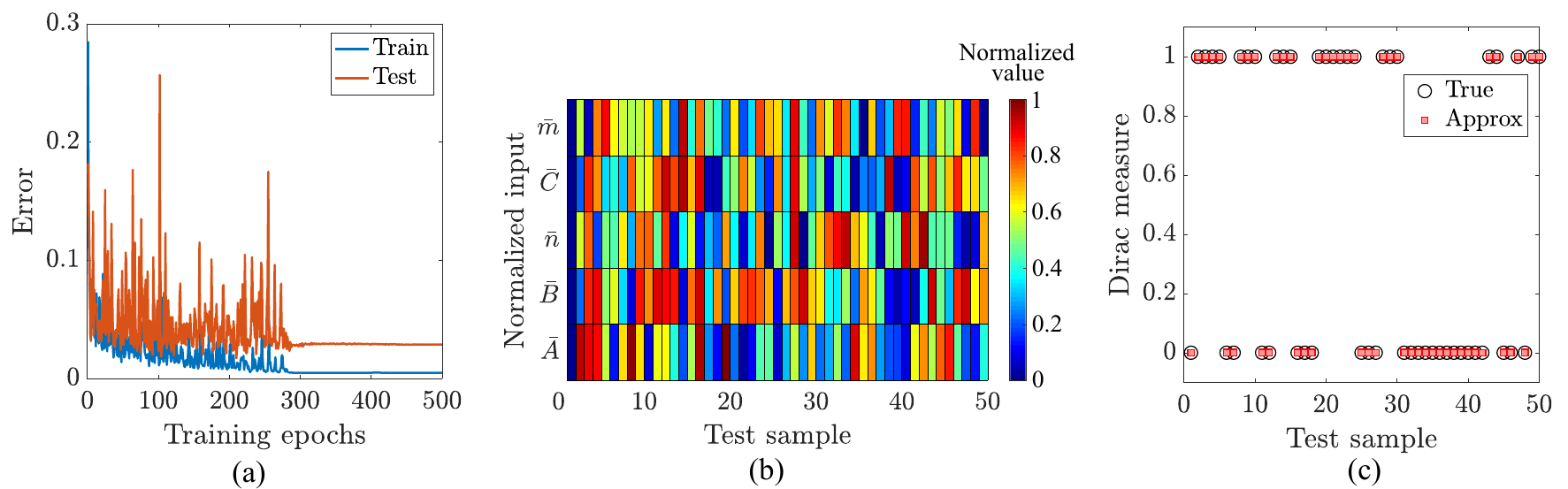}
\caption{Training and testing results for a typical performance indicator with threshold $Y_T = 1.03$~cm. (a) Convergence of training and testing errors. (b) Distribution of 50 randomly selected test samples. (c) Comparisons between true and approximate Dirac measures.}
\label{fig:data}
\end{figure}

Fig.~\ref{fig:data} shows the results of a typical learned performance indicator with threshold $Y_T = 1.03$~cm. Both training and testing error is included in Fig.~\ref{fig:data}(a) with minimum training and testing error to be $0.005$ and $0.0287$ respectively. A set of $50$ testing samples were randomly selected with the input $X = \{\hat{A},\hat{B},\hat{n},\hat{C},\hat{m}\}$ plotted in Fig. \ref{fig:data}(b). The resultant output $\delta_{F(X)}(\mathcal{Y}_c)$ is plotted in Fig.~\ref{fig:data}(c). It is evident from Fig. \ref{fig:data}(c) that the neural network prediction (marked in square) agrees very well with the true solution (marked in circle) regarding the cases of both $0$ and $1$.     

The computational cost of evaluating the performance indicator as shown in Eqn.~(\ref{eq:NN approx}) is shown in Table~\ref{tab:cost}. All calculations were preformed on a single core of Intel Skylake CPU
($2.1$~GHz) except the neural network training which was done on a NVIDIA P100 GPU with $3584$ CUDA cores. It is seen that evaluating $1$ sample of the performance indicator requires only $1.4$~ms which is $\mathcal{O}(10^5)$ faster than direct numerical simulation which is based on LS-DYNA. As a result, the computation time required to solve the optimization problems is only on the order of seconds.

\begin{table}[!ht] 
\centering
\caption{Computational cost (wall-clock time in seconds)}
\begin{tabular}{l l}
\hline
\hline
Method & ~\  Computation cost of the performance indicator\\
\hline
Direct numerical simulation (1 sample) &     ~~~~~~~~~~~~~~~~~~~~~~~~~~~~~$200$~~~ \\
Neural network surrogate 
 \begin{tabular}{@{}ll@{}}
                   Train (1500 samples) \\ Test (1 sample)\\
                 \end{tabular}  &  \begin{tabular}{@{}ll@{}}
                   ~~~~~~~~~~~~~~~~~~~~~~~~~~~~~40 \\  ~~~~~~~~~~~~~~~~~~~~~~~~~~~~~0.0014 \\
                 \end{tabular} \\
\hline
\end{tabular}
\label{tab:cost}
\end{table}

\subsection{Case studies} 
We are now in a position to demonstrate the power of our method by considering a series of case studies with varying levels of knowledge on input probabilities. We recall that such knowledge is applied as constraints in the optimization problems, Eqn.~(\ref{eq:optimization}). In this section, we will study multiple cases including mean and higher-order moment constraints, and complete and partial moment constraints. The corresponding constraints are also expressed as mathematical equations in our OUQ framework. In all cases, we have tested the penalty coefficients $\lambda_i,i=1,..,K,$ over multiple orders of magnitude ranging from $10$ to $10^5$. We finally fix all $\lambda_i$ at $10^3$ which provides an excellent compromise between computational speed and constraint satisfaction. The computation time required to solve the optimization problems is on the order of seconds.

\subsubsection*{Case 1: OUQ with mean constraints and comparison with other UQ methods}
As already mentioned, the present OUQ approach aims to predict the optimal bounds on the probability of failure (PoF). Therefore, it is both interesting and useful to verify the conservativeness and tightness of the obtained bounds. To this end, we first compare our learning-based OUQ with two other UQ methods, i.e., Monte Carlo (MC) sampling and concentration-of-measure (CoM) inequality.

The MC sampling requires explicit knowledge of the underlying probability measure. Therefore, we assume our input variable $X$  is distributed according to a multi-variable uniform measure $\mu(X)\sim U_n([0,1]^5)$. The MC sampling is performed over the ranges of the random parameters, using $1.2\times10^4$ samples with Latin Hypercube Sampling. Regarding CoM, we specifically employ the simple McDiarmid's inequality~\citep{sun2020rigorous, lucas2008rigorous} equipped with Genetic Algorithms, which requires only ranges of random inputs and hence supplies a working compromise between bound tightness and computational complexity. In the OUQ calculation, we assume that the only information we are given about the input random variables $X$ is their mean values $\mathbb{E}(X) \in \mathbb{R}^5$, such that the constraint function in Theorem~\ref{Winkler} is $f(X) = X$, with
\begin{equation}
\int_\mathcal{X}{X}{d\mu} = \mathbb{E}(X).
\end{equation}
The mean $\mathbb{E}(X)$ is then computed 
from the underlying uniform distributions $U_n$ used in the MC calculation.  

\begin{figure}[!ht]
\centering
\includegraphics[trim=0.7in 2.5in 0.7in 2.5in,clip,width=0.6\textwidth]{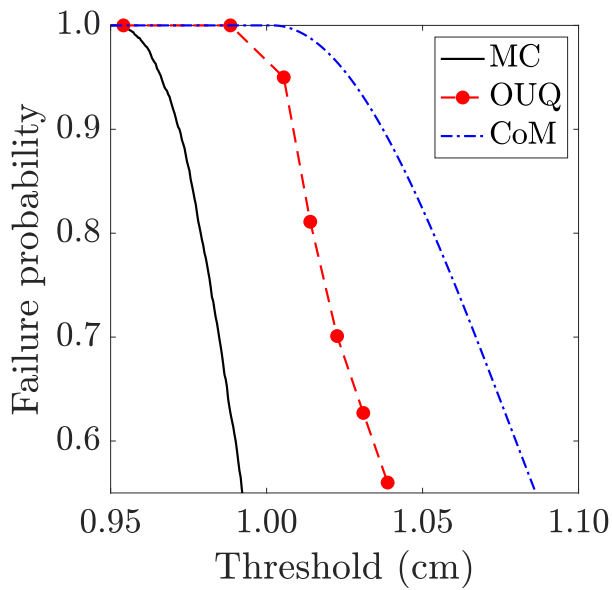}
\caption{Comparison of probability of failure computed from direct Monte Carlo (MC) sampling and Concentration of Measure (CoM) inequality.}
\label{fig:MC}
\end{figure}

Fig.~\ref{fig:MC} shows comparisons of the PoF calculated by MC and the two upper bounds calculated by CoM and OUQ. As expected, both the OUQ and CoM bounds lie uniformly above than the MC estimate, which illustrates the conservative character of the bounds and, by extension, of the corresponding designs. It is also noted that the OUQ bound is tighter than CoM bound. The reason is twofold. First, in addition to the ranges, our OUQ strategy makes use of mean constraints of the random inputs, which makes the probability description more accurate. Moreover, the upper bound obtained by OUQ is the optimal in the sense that further improvements inevitably require information about uncertainties other than or in addition to input ranges and mean values. 

\subsubsection*{Case 2: OUQ with higher-order moment constraints}
\begin{figure}[!ht]
\centering
\includegraphics[width=1.0\textwidth]{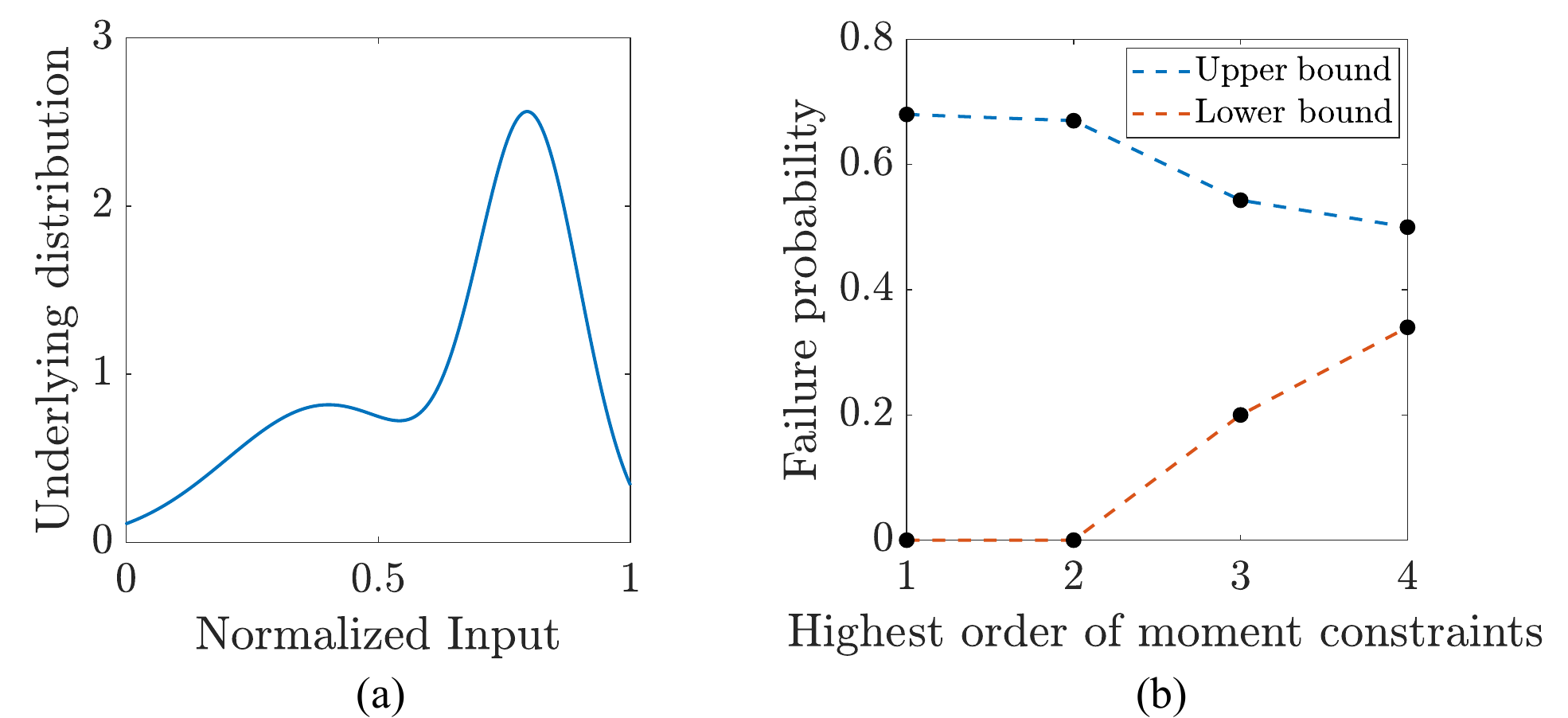}
\caption{Effects of highest-order of moment constraints. (a) Underlying bimodal distribution of random variables. (b) Convergence of bounds on the probability of failure.}
\label{fig:moment}
\end{figure}

We recall that our learning-based OUQ framework is capable of determining the optimal bounds on the PoF by leveraging all known information about uncertainties. Therefore, it is of particular interest to investigate the tightness of the bounds as the amount of known information increases. To this end, we consider the OUQ problem with increasing information on the input in the form of moment constraints. We start with an assumption that each component of our normalized input variables is identically and independently distributed from a truncated bimodal distribution $\nu([0,1])$ as shown in Fig.~\ref{fig:moment}. The true probability measure on $X$ is therefore $\mu = \nu \otimes \nu \otimes \nu \otimes \nu \otimes \nu$. In the UQ analysis, we assume that we are only able to access the moments of $\mu$ defined as: 
\begin{equation}
\mathbb{M}^j(X) = \int_{\mathcal{X}}X^jd\mu, \ \ \ \ j \in \mathbb{Z}^+,
\end{equation}
while the true distribution of $\mu$ is inaccessible. 

Fig.~\ref{fig:moment} (b) depicts the effects of adding the higher-order moment constraints on the optimal uncertainty bounds. We start with the case where the highest-order moment constraint is $\mathcal{A} = \{\mathbb{M}^1\}$, and proceed with adding an addition constraint $\mathbb{M}^2$ such that the inputs are now constrained by the set $\mathcal{A}=\{\mathbb{M}^1, \mathbb{M}^2\}$. We repeat this procedure until the highest moment constraint is $4$ with the corresponding constrained set $\mathcal{A}=\{\mathbb{M}^1,.., \mathbb{M}^4\}$. It is seen from the results that the upper bound $U(\mathcal{A})$ decreases with increasing moment constraints while the lower bound $L(\mathcal{A})$ increases. Importantly, the distance between $U(\mathcal{A})$ and $L(\mathcal{A})$ is seen to decrease by adding higher-order moment constraints. Moreover, one can observe that enforcing only the mean constraints and enforcing the first two-order moment constraints give almost the same bounds. However, adding the first three- and higher-order moment constraints will drastically reduce the space between the two bounds. It is expected as in the extreme case where all the moment constraints of $\mu(X)$ is enforced, there is no uncertainty in $\mu$ itself hence $U(\mathcal{A})$ and $L(\mathcal{A})$ should be equal and both should be equivalent to $\mathbb{P}[Y\in \mathcal{Y}_c]$. Therefore, considering more information about uncertainties will help enhancing the tightness of the OUQ bounds. 

\subsubsection*{Case 3: OUQ with partial moment constraints}
\begin{figure}[!ht]
\centering
\includegraphics[width=1.0\textwidth]{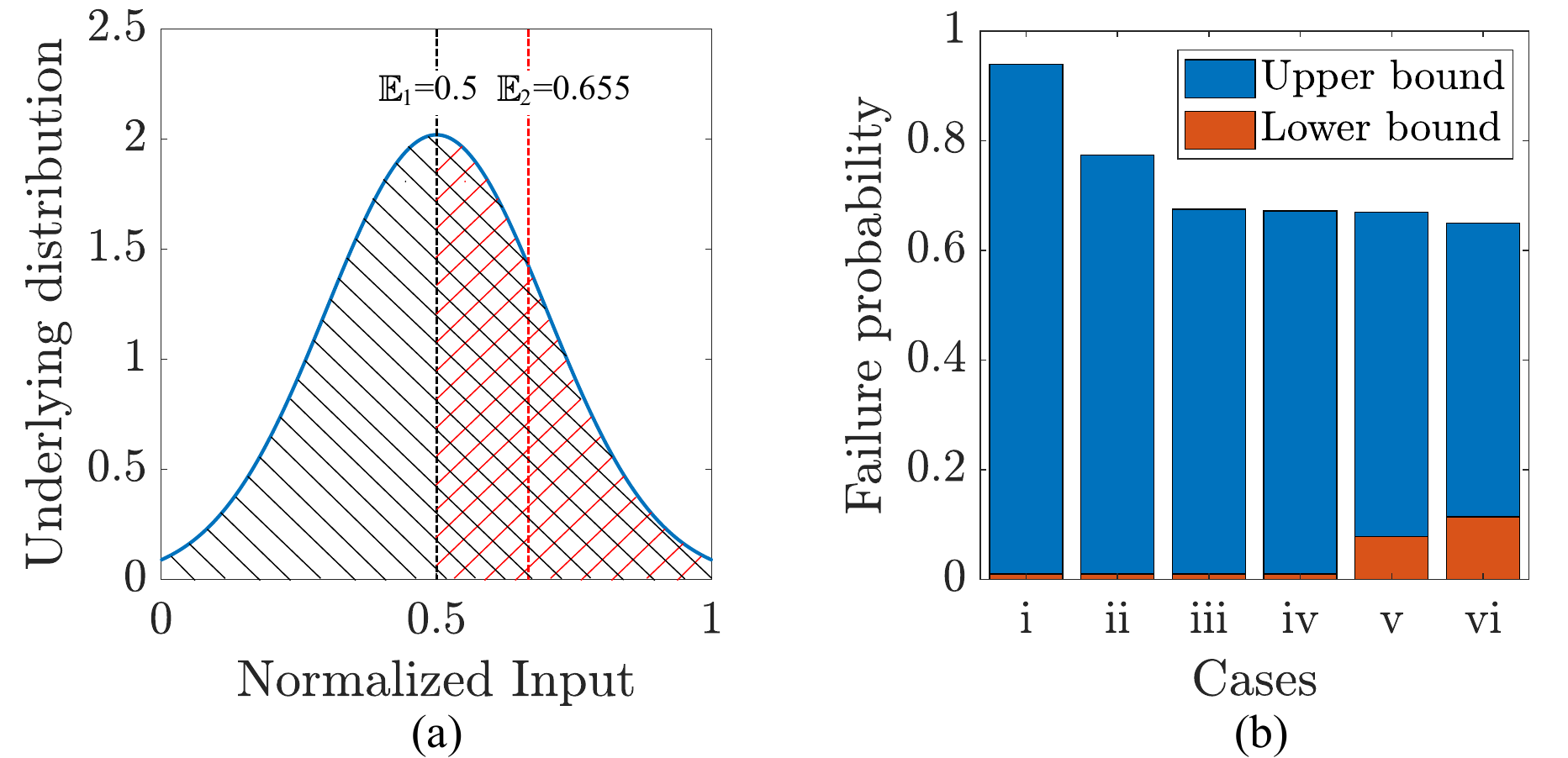}
\caption{Effects of first-order moment constraints in subintervals. (a) Schematic illustration of mean constraints on the interval $[0,1]$ and subinterval $[0.5,1]$ of random variables. (b) Convergence of bounds on failure probability. Case i: no partial constraint. Case ii: partial constraint on $\{\bar{A}\}$. Case iii: partial constraint on $\{\bar{A}, \bar{B}\}$. Case iv: partial constraint on $\{\bar{A}, \bar{B}, \bar{n}\}$. Case v: partial constraint on $\{\bar{A}, \bar{B}, \bar{n}, \bar{C}\}$. Case vi: partial constraint on $\{\bar{A}, \bar{B}, \bar{n}, \bar{C}, \bar{m}\}$.}
\label{fig:gaussian}
\end{figure}

It is also notable that the moment constraints enforced in the optimization do not need to be restricted to the entire space of the random distribution. We can also harness the information that characterizes the statistics over some subspaces of the inputs. This corresponds to the situation where the input probability measure is only known on a coarse topology, and we shall refer to such type of information as partial constraints. As an example, we consider the case where the underlying  probability measure of our input random variable is given by the product of 5 truncated Gaussian measure $\mathcal{N}([0,1])$. We are only able to access the mean of our input random variable $\mathbb{E}(X) = \mathbb{E}_1\otimes\mathbb{E}_1\otimes\mathbb{E}_1\otimes\mathbb{E}_1\otimes\mathbb{E}_1$ over the whole domain $X \in [0,1]^5$, but are given additional information regarding the mean of inputs supported on the subdomain $X \in [0.5, 1]^5$ with $\mathbb{E}(X) = \mathbb{E}_2\otimes\mathbb{E}_2\otimes\mathbb{E}_2\otimes\mathbb{E}_2\otimes\mathbb{E}_2$, as illustrated in Fig.~\ref{fig:gaussian}(a). As a result, we have $6$ cases in this test: Case i: no partial constraint; Case ii: partial constraint on $\{\bar{A}\}$; Case iii: partial constraint on $\{\bar{A}, \bar{B}\}$; Case iv: partial constraint on $\{\bar{A}, \bar{B}, \bar{n}\}$; Case v: partial constraint on $\{\bar{A}, \bar{B}, \bar{n}, \bar{C}\}$; and Case vi: partial constraint on $\{\bar{A}, \bar{B}, \bar{n}, \bar{C}, \bar{m}\}$. The corresponding optimal upper and lower bounds are shown in Fig.~\ref{fig:gaussian}(b). We observe that the upper bound decreases with increasing partial constraints, while the lower bound increase with increasing partial constraints. It is also notable that the upper bound is most sensitive to the partial information enforced on $\bar{A}$ and $\bar{B}$ while the lower bound is most sensitive to that enforced on $\bar{C}$ and $\bar{m}$.   

\subsection{Performance certification and safety design}

\begin{figure}[!ht]
\centering
\includegraphics[width=1.0\textwidth]{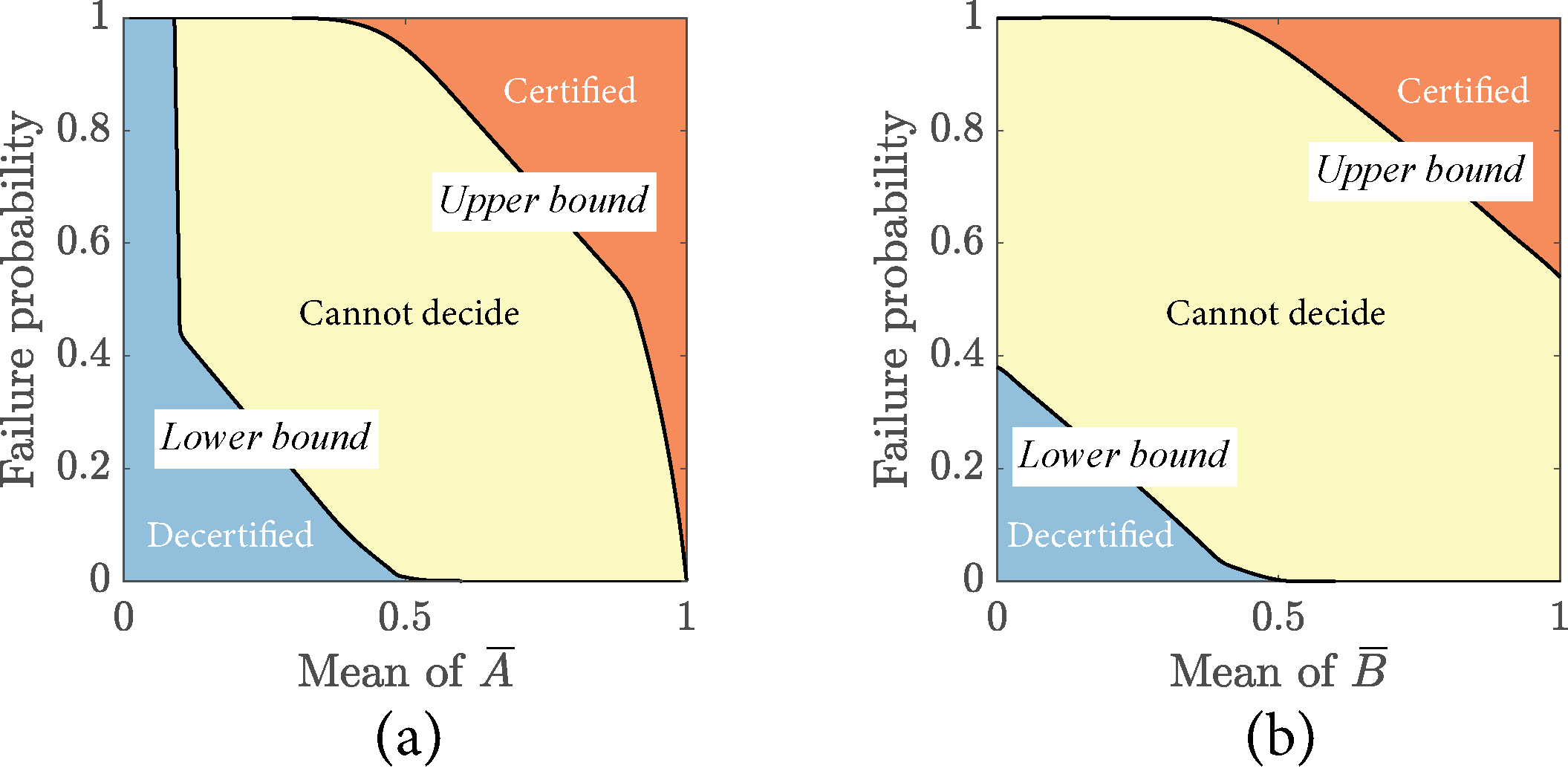}
\caption{Certification and design under uncertainties. (a) Mean of $\{
\bar{A}\}$ as the design parameter. (b) Mean of $\{\bar{B}\}$ as the design parameter.}
\label{fig:mean}
\end{figure}

We shall end our numerical studies with a demonstration of how our OUQ method can be used in performance certification and safety design. By way of example, we consider again the plate/projectile system analyzed in the foregoing. We follow the materials-by-design strategy and seek to ``design" a material with desired properties that can meet the prespecified design requirements. Those material properties can be uncertain with unknown probability distributions that are uncontrollable from manufacturing. Therefore, it is impractical to determine a specific value or probability measure for them. Instead, their statistical moments, e.g., mean or variance, can be easily accessible and adjusted in material fabrication using a small amount of data. In this regard, we choose our design parameters to be the mean values of two Johnson-Cook parameters that contribute most significantly to the upper bound on PoF, i.e., the normalized yield stress $\bar{A}$ and the normalized strain-hardening modulus $\bar{B}$. We seek to determine the minimum values of these two material properties that guarantee the given design tolerance and performance threshold.  

The dependence of the upper and lower PoF bounds on the means of $\bar{A}$ and $\bar{B}$ is shown in Fig.~\ref{fig:mean}. These bounds are determined by solving the OUQ optimization at different mean values of $\bar{A}$ or $\bar{B}$ while constraining the other four normalized Johnson-Cook parameters in the corresponding ranges with a fixed mean of $0.5$. The threshold of the maximum backface deflection is also fixed at $1$~cm. As expected, for both $\bar{A}$ and $\bar{B}$, the PoF bounds decrease with the increase of means of $\bar{A}$ and $\bar{B}$, since the plate is more likely to be stronger and stiffer. It is also notable in Fig.~\ref{fig:mean}(a) that when the mean of $\bar{A}$ is less than 0.09 (i.e., $A = 205$ MPa), both the upper and lower bounds on the PoF converge to $1$. This means that the system is unsafe with $100\%$ confidence regardless of the values of other material properties as long as they satisfy the constraints. Therefore, under this scenario the impact performance of the plate is governed by the yield stress of AZ31B Mg alloys. Moreover, each subfigure of Fig.~\ref{fig:mean} is divided by the lower and upper bounds into three regions labeled as ``Decertified", ``Cannot decide" and ``Certified". As a result, for a given tolerance $\epsilon$ on the PoF, these regions provide reference ranges of the design parameters to achieve the rigorous certification criterion illustrated in Fig.~\ref{fig:certify}. That is, the system is provable safe if the design parameter and the PoF tolerance are located in the region ``Certified", is provable unsafe if in the region ``Decertified", and the safety of the system cannot be decided if in the region ``Cannot decide".

\section{Concluding remarks} 
\label{sec:summary}

We have presented a learning-based UQ framework that enables the computation of the optimal (supremum and infimum) bounds on the probability of failure (PoF). This framework is based on the Optimal Uncertainty Quantification methodology, which does not require explicit knowledge or presumptions on the input/prior distributions. Rather, we only make use of the available information that can be partial or imperfect on the input variables (e.g., mean, variance, or other higher-order moments). Such information is then converted as constraints to an optimization problem where we search for the extremal input (prior) distributions which maximize/minimize the PoF. The optimization problem is then solved with the aid of machine learning, where neural networks are used as surrogate models for computing the performance indicator. Using the obtained upper and lower PoF bounds, rigorous performance certification and safety design can be conducted for a given design threshold and failure tolerance. We have demonstrated the capability of our approach in problems involving the ballistic impact of magnesium (Mg) alloy plates where the uncertainties arise from the constitutive model parameters.

Several significant findings afforded by the calculations are noteworthy. The learned neural network for performance indicator is extremely computationally efficient. It is $\mathcal{O}(10^5)$ faster than the finite-element based direct numerical simulations while maintaining high approximating accuracy with errors less than $3\%$. As a result, the computation time required to solve the optimization problems is only on the order of seconds. Moreover, the upper and lower bounds obtained by our framework are tighter than those calculated by the Concentration of Measure inequality, which has simple theoretical formulations but supplies non-optimal PoF bounds. We have also verified the conservativeness of our OUQ bounds, and the attendant tightness can be significantly increased by either adding higher-order moment constraints over the entire input spaces or considering partial moment constraints over some subspaces. As a result, we have found that the ballistic performance of the plate under consideration is most sensitive to the yield stress $A$ and the strain-hardening modulus $B$ of the AZ31B Mg alloys, since the PoF upper bound decreases drastically with the increase of moment constraints on them.

Our learning-based framework employs imperfect or partial knowledge about uncertainties. Therefore, the information set of uncertainties $\mathcal{A}$ in Eqn.~(\ref{eq:info}) is open to a great deal of generalization. In principle, any information about the response function $F$ and the probability measure $\mathbb{P}$ can be harnessed to define a set of admissible scenarios $\mathcal{A}$ for the optimization problems in Eqn.~(\ref{eq:bounds}). This paper makes use only of the equality constraints of moments on the inputs. In contrast to most UQ approaches, our approach do not require the input variables $X_i, i = 1,...,m$, to be independently distributed. The information about correlations of the input random variables can be included in the definition of $\mathcal{A}$. Other types of constraints can also be considered in $\mathcal{A}$, such as generalized moment constraints on outputs and generalized moment constraints on input domain partitioning. In addition to equality constraints, inequality constraints of aforementioned types can also be implemented in $\mathcal{A}$. We note that although increasing the amount of additional information/constraints on input probability measure results in tighter OUQ bounds, it makes OUQ optimization problems more challenging as the dimension of the optimization task grows linearly with the number of constraints. In the current study, we have only considered partial information on input probability measures in the form of moment constraints. The task of considering other type of information and constraints remains unexplored and we shall leave it for future studies.

\section{Acknowledgement} 
The authors are grateful to Profs. Kaushik Bhattacharya and Michael Ortiz, for helpful discussions. XS gratefully acknowledges the support of the University of Kentucky through the faculty startup fund. BGL gratefully acknowledges the support of Granta Design through the startup fund.  
\bibliographystyle{abbrvnat}
\bibliography{references}

\end{document}